\documentclass[sigconf,screen]{acmart}
\usepackage{amsmath}
\usepackage{amsfonts}
\usepackage{algorithm}
\usepackage{algpseudocode}
\usepackage{bm}
\usepackage{multirow}
\usepackage{graphicx}
\usepackage{caption}
\usepackage[font=footnotesize,labelfont=rm,textfont=rm]{subcaption}
\usepackage{float}
\usepackage{microtype}
\usepackage{balance}

\AtBeginDocument{%
  \providecommand\BibTeX{{%
    \normalfont B\kern-0.5em{\scshape i\kern-0.25em b}\kern-0.8em\TeX}}}

\setcopyright{rightsretained}
\acmPrice{}
\acmDOI{10.1145/3611643.3613896}
\acmYear{2023}
\copyrightyear{2023}
\acmSubmissionID{fse23industry-p96-p}
\acmISBN{979-8-4007-0327-0/23/12}
\acmConference[ESEC/FSE '23]{Proceedings of the 31st ACM Joint European Software Engineering Conference and Symposium on the Foundations of Software Engineering}{December 3--9, 2023}{San Francisco, CA, USA}
\acmBooktitle{Proceedings of the 31st ACM Joint European Software Engineering Conference and Symposium on the Foundations of Software Engineering (ESEC/FSE '23), December 3--9, 2023, San Francisco, CA, USA}
\received{2023-05-18}
\received[accepted]{2023-07-31}

\begin{document}





\title{Beyond Sharing: Conflict-Aware Multivariate Time Series Anomaly Detection}


\author{Haotian Si}
\authornote{Also with University of Chinese Academy of Sciences, Beijing, China.}
\affiliation{%
  \institution{Computer Network Information Center, Chinese Academy of Sciences}
  \city{Beijing}
  \country{China}}

\author{Changhua Pei}
\authornote{Changhua Pei is the corresponding author. Email: chpei@cnic.cn}
\affiliation{%
  \institution{Computer Network Information Center, Chinese Academy of Sciences}
  \city{Beijing}
  \country{China}}

\author{Zhihan Li}
\affiliation{%
  \institution{Kuaishou Technology}
  \city{Beijing}
  \country{China}}

\author{Yadong Zhao}
\authornotemark[1]
\affiliation{%
  \institution{Computer Network Information Center, Chinese Academy of Sciences}
  \city{Beijing}
  \country{China}}

\author{Jingjing Li}
\author{Haiming Zhang}
\affiliation{%
  \institution{Computer Network Information Center, Chinese Academy of Sciences}
  \city{Beijing}
  \country{China}}

\author{Zulong Diao}
\authornote{Also with Purple Mountain Laboratories, China.}
\affiliation{%
  \institution{Institute of Computing Technology, Chinese Academy of Sciences}
  \city{Beijing}
  \country{China}}

\author{Jianhui Li}
\affiliation{%
  \institution{Computer Network Information Center, Chinese Academy of Sciences}
  \city{Beijing}
  \country{China}}

\author{Gaogang Xie}
\affiliation{%
  \institution{Computer Network Information Center, Chinese Academy of Sciences}
  \city{Beijing}
  \country{China}}

\author{Dan Pei}
\affiliation{%
  \institution{Tsinghua University}
  \city{Beijing}
  \country{China}}

\renewcommand{\shortauthors}{Haotian Si, Changhua Pei, Zhihan Li, \\ Yadong Zhao, Jingjing Li, Haiming Zhang, Zulong Diao, Jianhui Li, Gaogang Xie, Dan Pei}
\begin{abstract}
Massive key performance indicators (KPIs) are monitored as multivariate time series data (MTS) to ensure the reliability of the software applications and service system. 
Accurately detecting the abnormality of MTS is very critical for subsequent fault elimination. 
The scarcity of anomalies and manual labeling has led to the development of various self-supervised MTS anomaly detection (AD) methods, which 
optimize an overall objective/loss encompassing all metrics' regression objectives/losses. However, our empirical study uncovers the prevalence of conflicts among metrics' regression objectives, causing MTS models to grapple with different losses.
This critical aspect significantly impacts detection performance but has been overlooked in existing approaches.
To address this problem, by mimicking the design of multi-gate mixture-of-experts (MMoE), we introduce \textbf{CAD}, a \textbf{C}onflict-aware multivariate KPI \textbf{A}nomaly \textbf{D}etection algorithm. CAD offers an exclusive structure for each metric to mitigate potential conflicts while fostering inter-metric promotions. 
Upon thorough investigation, we find that the poor performance of vanilla MMoE mainly comes from the \textbf{input-output misalignment} settings of MTS formulation and \textbf{convergence issues} arising from expansive tasks. To address these challenges, we propose a straightforward yet effective task-oriented metric selection and p\&s (personalized and shared) gating mechanism, which establishes CAD as the first practicable multi-task learning (MTL) based MTS AD model.
Evaluations on multiple public datasets reveal that CAD obtains an average F1-score of 0.943 across three public datasets, notably outperforming state-of-the-art methods. 
Our code is accessible at \textbf{\url{https://github.com/dawnvince/MTS_CAD}}.

\end{abstract}



\maketitle

\section{Introduction}
\label{sec: introd}
With the rapidly increasing number of Internet applications and the number of users, ensuring the stability of Internet services and software systems is now very important. Service Level Agreements (SLAs) on service reliability ask for real-time and accurate incident identifications. To improve the observability of the system, operators deploy monitoring programs to produce a large amount of time series data to monitor the status (\textit{i.e.}, metrics) of the entities (\textit{e.g.}, software systems, online services) in different application domains (\textit{e.g.}, IT systems, manufacturing industry), providing rich information for anomaly detection and incident alert. Traditional metric-level anomaly detection methods determine whether to report an incident or not based on manually setting thresholds on each metric, becoming unqualified for more strict SLAs and less effective due to the explosive growth in the number of metrics.



There is a natural tendency for anomalies to be detected in univariate time series (UTS)~\cite{uni1, uni2, uni3}, that is to say, the anomaly detection is performed on each metric separately. However, researchers have found that in the complex real-world system, different UTS interact with each other, which is called \textit{inter-metric dependency}~\cite{interfusion}. Anomaly detection on 
each metric separately may lead to considerable false negatives. To better illustrate this, we take CPU utilization of certain services and Query Per Second (QPS) metrics as examples. At the time $t$, we observe that QPS drops while CPU utilization increases. Under normal circumstances, QPS has a positive relationship with the CPU. It indicates an anomaly because the normal inter-metric dependency is violated. The UTS-based anomaly detection model fails to detect the anomaly as each metric's absolute value is within the normal range. On the contrary, this anomaly can be easily detected by the MTS anomaly detection model as it can not only model the intra-metric temporal dependency but also models the inter-metric dependency. In this paper, we focus on anomaly detection for multivariate time series data (MTS for short hereafter).  
Various MTS anomaly detection methods are proposed to model the correlations between different metrics. They almost all use self-supervised learning frameworks, more specifically, regression learning due to the scarcity of anomalies and manual labeling. All metrics' regression objectives compose the overall optimization objective/loss. 

After considerable empirical investigation, we observe that the objectives of different subtasks may not be consistent, and can even be disparate in some cases. A representative case is that while most metrics have stable baselines and patterns, on one single metric or some groups of metrics, either baseline drifts or inherent stochastic fluctuations (BD\&ISF for short) happen frequently, both of which are not labeled as anomalies, which is contrary to our naive instincts. Analysis and consultation with relevant people indicate the justification for the existence of these metrics in the real-world scenario. In this case, the objectives of stable metrics induce the model to pay more attention to subtle variation (hence sensitive to BD\&ISF), while metrics with BD\&ISF require the model to be insensitive to these changes. We refer to this as \textbf{conflicts}. As a result, the objectives at odds would cause the gradients of model parameters to descend in different or even opposite directions. This weakens the ability of the model and ultimately results in a decrease in detection performance.

However, We find that existing models cannot deal with the loss conflicts among metrics. These methods can be roughly divided into two classes: graph-based and sequence-based. The graph-based methods~\cite{deng2021graph, dvgcrn, fusagn} take each metric as a node, construct the complete graph between them and apply extensive techniques of graph neural networks to model inter-metric dependency. In sequence-based methods, attention-based~\cite{tranad, atten} or RNN-based~\cite{interfusion, lstmndt} mechanisms are widely used to extract sequential information. Unfortunately, none of them thinks of conflicts between learning objectives from a framework perspective and takes active action to isolate the effect scopes of these contradictory objectives. 
This causes poor performance when conflicts occur. 

The aforementioned findings call for a detection method that is capable of eliminating conflicts. As the conflicts are discrepant objectives essentially, we expect that the gradient descent process of conflicting losses can be isolated to some extent. A naive idea is to train a separate model for each metric that takes into consideration the influences of other metrics. However, it is unrealistic to train one model using all data for each metric since the number of metrics to be detected is increasing rapidly. Google~\cite{MMOE} has proposed multi-gate mixture-of-expert (MMoE), using a group of experts and multiple gates to leverage the correlations while avoiding interference among tasks under a multi-task learning framework. We prepare an exclusive structure for each metric and attempt to borrow ideas from MMoE in the multivariate time series anomaly detection domain.
Unfortunately, unlike the information retrieval domain that MMoE is designed for, 
introducing the idea of MMoE in the anomaly detection domain cannot meet our expectations due to the following challenges:

\begin{itemize}
    \item \textbf{Misalignment of input and output spaces}: Due to the employment of self-supervised regression learning or its variants, the output of an isolated subtask (one specific metric) is a sub-space of the whole input feature space (all metrics). 
    This requires an appropriate mapping mechanism for subtasks to extract more related information from the large feature spaces.
    \item \textbf{Convergence issues}: A monitor system would produce massive metrics for anomaly detection, each of which is viewed as a subtask. The number of subtasks in MTS tasks is dozens of times more than that in the vanilla MMoE scenario. As the number of subtasks increases, the naive gate structure fails to spread the gradients to experts in a stable way, causing oscillations in the parameter updates of the experts and finally influencing the convergence of the model.
\end{itemize}

To address the above challenges, in this paper, we propose \textbf{CAD}, a \textbf{C}onflict-aware multivariate \textbf{A}nomaly \textbf{D}etection algorithm. CAD trains a number of experts which can model the temporal-spatial dependency from multiple perspectives with the help of convolution networks. Then the expert networks are combined together in a weighted summing manner. For one specific metric, an automatic gating mechanism can assign personalized weights to different experts. In this way, compared with current MTS anomaly detection models which share the same network, CAD can flexibly learn personalized inter-metric dependencies for each target detection metric while isolating the negative effect brought by conflicts. 

To handle the misalignment of input and output space of MTS, a task-oriented feature selection as well as a p\&s (personalized and shared) gating mechanism is designed. Additionally, the p\&s gating mechanism greatly improves the robustness and convergence of the model as the number of tasks increases. The personalized gate selects the most related experts for each task which can prevent the collapse of experts from learning the same thing. The shared gate ensures robust expert selection to make the expert network converge quickly.



The main contributions of our paper can be summarized as follows:
\begin{itemize}
    \item Our work is based on an observation that has never been considered before that the discordance of data distribution is likely to cause conflicts between the objectives of metrics, which do harm to the detection performance of existing methods. We propose \textbf{CAD}, a \textbf{C}onflict-aware MTS \textbf{A}nomaly \textbf{D}etection algorithm, to address the limitations of existing models when dealing with conflicts in MTS.
    \item We summarize the key challenges encountered when eliminating the impact of conflicts, proposing a task-oriented feature selection to prompt the subtasks to focus more on their own patterns and a p\&s gate mechanism to make the model more robust when facing massive subtasks. Well-designed experts are capable to extract both temporal and inter-metric dependency embedded in MTS in multiple perspectives, significantly enriching the expressivity of the whole model.
    \item We conduct comprehensive evaluations on multiple open-source public datasets, showing that CAD outperforms the state-of-the-art by considerable margins (average 4.3\% to 37.9\% improvement over three datasets under best-F1 using point-adjustment approach and 4.2\% to 93.1\% under best-F1 using the k-th point-adjustment approach). Our code is publicly released. 
\end{itemize}

    

This paper is organized as follows. In Section~\ref{sec:preliminaries}, We give a formulation of MTS AD tasks and an illustration of conflict. The methodology of our model is presented in Section~\ref{sec:model} and the experimental setup is presented in Section~\ref{sec:settings}. We offer comprehensive evaluations of our model in Section~\ref{sec:evaluation}. We review the related works in Section~\ref{sec:related work} and conclude our paper in Section~\ref{sec:conclusion}.


%


\section{Background}
\label{sec:preliminaries}
In this section, we formulate the MTS anomaly detection problem and provide a brief introduction to MMoE. Then we use a case 
to help readers better understand the concept of conflict.
\subsection{Problem Formulation}
\label{subsec:wwt}
Multivariate time series consists of multiple curves that are time-aligned, each of which represents successive observations of a metric over a long period of time. MTS  with $T$ consecutive timestamps and $K$ metrics is represented by an ordered sequence:
$$\mathbf{X} = 
\begin{Bmatrix}
\mathbf{x_1}, \cdots, \mathbf{x_i}, \cdots, \mathbf{x_T}
\end{Bmatrix},
$$
where $\mathbf{x_i}$ is defined as the set of observations for all metrics at the particular timestamp $i \in [1, T]$:
$$\mathbf{x_i} = 
\begin{pmatrix}
x_i^1, x_i^2, \cdots, x_i^K
\end{pmatrix}^\top,
$$
as each datapoint $x_i^j$ denotes the value of metric $j$ at this time point.

Given current observation $\mathbf{x_t}$ at timestamp $t$ on the test set, a MTS anomaly detection system calculates the anomaly score in some way based on historical observations, then identifies whether $\mathbf{x_i}$ is anomalous or not by comparing its anomaly score with threshold.

In MTS anomaly detection task, contextual observations, i.e., observations at nearby timestamps, play an essential role in understanding current data because they notably describe the relevant temporal patterns~\cite{omni, usad}. Thus for observation $\mathbf{x_t}$, rather than simply using standalone vector $\mathbf{x_{t-1}}$, we take a sliding window of length $l$, $
\begin{Bmatrix}
\mathbf{x_{t-}}_l, \cdots, \mathbf{x_{t-1}}
\end{Bmatrix}$ (denoted by $\mathbf{w_t}$), as input to precisely capture sequential dependency in our implementation.

\subsection{Basis of MMoE framework}
The MoE layer proposed by Eigen et.al.~\cite{moe1} and Shazeer et.al.~\cite{moe2} leverages the techniques of ensemble learning, introducing a gate module to make the network more sparse. This allows for the incorporation of more parameters without incurring additional computational costs. Ma et.al.~\cite{MMOE} further adapt the layer to multi-task learning, substituting the only gate with multi-gates each of which is allocated to a specific task individually, to utilize shared embedding while handling complex task correlation issues gracefully. The k-th task of the Multi-gate Mixture-of-Experts (MMoE) Model can be formulated as follows:
$$
y_k = h_k(g_k(x) \odot F(x))
$$
Here, $\odot$ denotes element-wise product, $F(*)$ denotes the set of intermediate results produced by experts, $g(*)$ denotes gate structure and $h_k(*)$ denotes the model of the k-th downstream task, which is known as \textbf{tower} network. Given $M$ experts' output embeddings $e_1, \cdots, e_M$ , $g_k \in \mathbb{R}^{1\times M}$ includes the weight of experts, hence $g_k(x) \odot F(x)$ assembles all embeddings into one as the input of k-th tower. Each tower would give final results based on its particular aggregation of embeddings.

\begin{figure}[tb]
    \centering
    \includegraphics[width=0.4\textwidth]{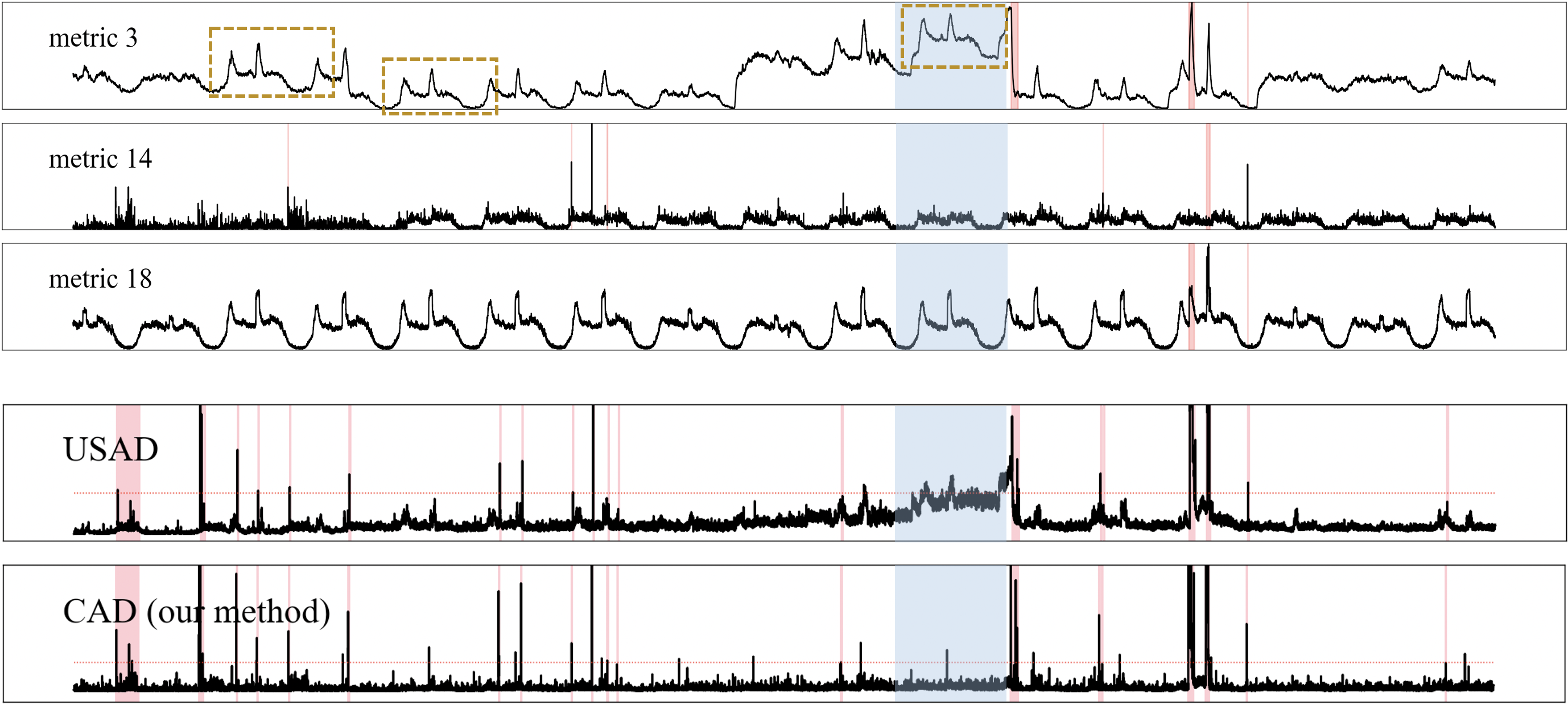}
    \caption{Illustration of one possible conflict among metrics. Anomalous segments at the metric level are highlighted in red, and the segment corresponding to reasonable drift in metric 3 is highlighted in blue. The regions enclosed by the yellow boxes share similar distribution. A previous method's detection result as well as ours is zoomed in and listed below. Red backgrounds in the results are the union set of all metrics' anomalies (including the three metrics plotted).}
    \label{fig:case}
\end{figure}

\begin{figure*}[tbp]
    \centering
    \includegraphics[width=0.85\textwidth]{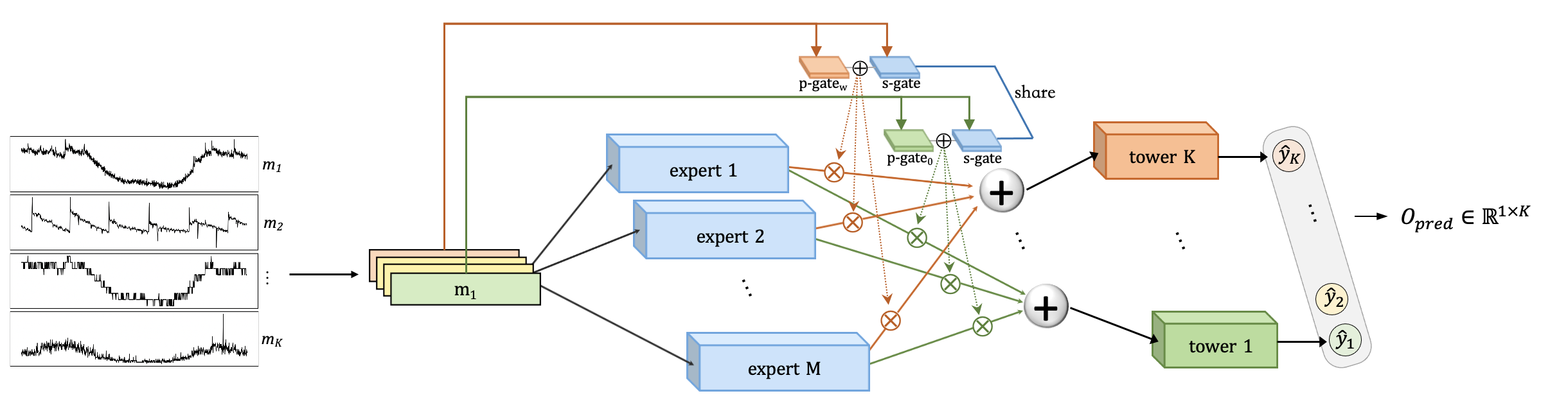}
    \caption{Network Architecture of CAD.}
    \label{Fig.model}
\end{figure*}

\subsection{Illustration example of conflicts}
An illustrative example with regard to conflicts is presented to motivate our approach (Fig.~\ref{fig:case}). In a dataset sampled from a real-world software system~\cite{omni}, there are three metrics: metric 3, metric 14 and metric 18. During one particular period, sudden baseline drift (highlighted in blue in Fig.~\ref{fig:case}) happens in metric 3, while all other metrics behave normally. Despite the drift, the distribution of data in this period is very similar to two previous normal distributions (enclosed by the yellow boxes). Meanwhile, the baselines of all three regions are at different levels, showing the metric's inherent sensitivity to normal external factors, e.g., load balancing strategy. Thus we consider that no anomalies occurred in this period which is confirmed by labels (highlighted in red in Fig.~\ref{fig:case}).
While this change is reasonable, the baseline drift of metric 3 misleads the parameter updating of existing models. An obvious drift can be observed in the output anomaly score of a previous model (shown as USAD in Fig.~\ref{fig:case}), which misjudges this period as an anomaly eventually.

The detection result indicates that our method could handle conflicts effectively while other previous works perform poorly as they are influenced by conflicts to some extent. More detailed discussions and evaluations on this case can be found in Sec.~\ref{sec:evaluation}.

\section{Methodology}
\label{sec:model}
In this section, we first give an overview of the system. Then we introduce the sub-components of the model individually, including task-oriented feature selection, expert network, personalized \& shared gate and tower network. At last, we introduce the loss balance module.
\subsection{Overall Structure}
As illustrated in Fig. \ref{Fig.model}, we leverage a group of task-oriented isolated structures to tackle the conflict problems discussed in Section~\ref{sec: introd}. Since previous works have no independent structures to determine whether inter-metric dependencies are helpful or harmful to a specific metric, the correlation may have a negative impact on describing reasonable patterns of the current metric. Our approach, as each task has its own structure to autonomously select various expert-derived features that are highly correlated with it, solves the conflicts in an elegant manner. Moreover, the isolated tower becomes even more focused on its own metric since irrelevant information is separated out.

However, original frameworks like MMoE are unacclimatized when facing MTS issues. Firstly, \textbf{misalignment of input and output spaces} distract the gates from focusing on more relevant patterns. Secondly, well-designed structures are required to capture the complicated temporal-spatial dependency. Thirdly, massive metrics bring more interference to feature extraction, which needs a way for experts to \textbf{converge on specific traits}.

To address the problems mentioned above, we design CAD, a hierarchical unsupervised and forecasting-based model.
Then a two-dimensional input window is fed into multiple expert networks including a convolution layer to enrich the representation of temporal and inter-metric dependency (Section~\ref{subsec: expert}). The hybrid gating mechanism (Section~\ref{subsec: psgate}), which is designed for massive metrics, transforms the feature space selected for each metric (Section~\ref{subsec: feature select}) into a set of weights to fuse embeddings extracted by different experts. The embedding is then sent to its corresponding tower network (Section~\ref{subsec: tower}). All towers jointly finalize anomaly scores based on the gap between predictions and ground-truth values. For the multiple losses balance problem, we preprocess the data to balance all metrics' losses (Section~\ref{subsec: balance}). These components are then described in detail.

\subsection{Task-oriented Feature Selection}
\label{subsec: feature select}
In a typical scenario where MMoE works, there is no obvious correlation between the downstream task and the input data. In other words, each dimension of the input data is equally significant for the current task, thus each gate calculates the weight combination of experts based on all input data. This assumption, however, does not hold true in MTS anomaly detection. In this case, for a subtask determining if the particular metric is abnormal, it is counterintuitive that historical information in \textbf{its own feature space} is as important as the one in \textbf{overall input feature spaces} composed of other metrics. More metrics mean more possibility to distract the limited volume structure, especially when conflicts and reasonable drifts exist among metrics (discussed in Section~\ref{sec: drift}). We split the input at metric level, using its own \textbf{local} time window instead of \textbf{all} metrics' windows to prompt gate structure for learning personalized mappings from time series to distribution over experts. 
Assuming that $\mathbf{w_t^k}$ denotes time window of the k-th metric in $\mathbf{w_t}$ defined in Sec.~\ref{subsec:wwt}, the embeddings of raw input $E$ at time $t$ consist of all experts' outputs:
\begin{equation}
    E^{(t)}(\mathbf{w_t}) = [f_1(\mathbf{w_t}), f_2(\mathbf{w_t}), \cdots, f_M(\mathbf{w_t})]
\end{equation}
where $f_i(x)$ denotes the output of the i-th expert network given input $x$, which is described in detail in Sec.~\ref{subsec: expert}. Then the embeddings $B^{(t,k)}$ sent into subtask $k$ are combined according to weights given by gate $G_k$:
\begin{equation}
    B^{(t,k)} = G_k(\mathbf{w_t^k}) \odot E^{(t)}(\mathbf{w_t})
\end{equation}

Finally, each subtask $k$ corresponding to the metric $k$ calculates the prediction value $\hat{y}_k^{(t)}$:
\begin{equation}
    \hat{y}_k^{(t)} = Tower_k(B^{(t,k)})
\end{equation}

\subsection{Expert Network}
\label{subsec: expert}

Expert networks are the main structure for extracting ample traits in time series. Each expert consists of one convolution layer and two feed-forward layers. Previous works primarily use RNN-based units, e.g. LSTM or GRU, to capture temporal patterns~\cite{omni, madgan, dvgcrn}. Whereas, its inherently sequential nature makes it impossible for parallel computation, dramatically increasing training time. By contrast, convolution network, known as a structure with superior computational parallelism, shows its powerful capability of extracting features in time series tasks~\cite{tcn, lstnet, dsanet}. Especially, kernels can sharpen the change within a successive region, which suits MTS anomaly detection tasks even better. 

\begin{figure}[htb]
    \centering
    \includegraphics[width=0.4\textwidth]{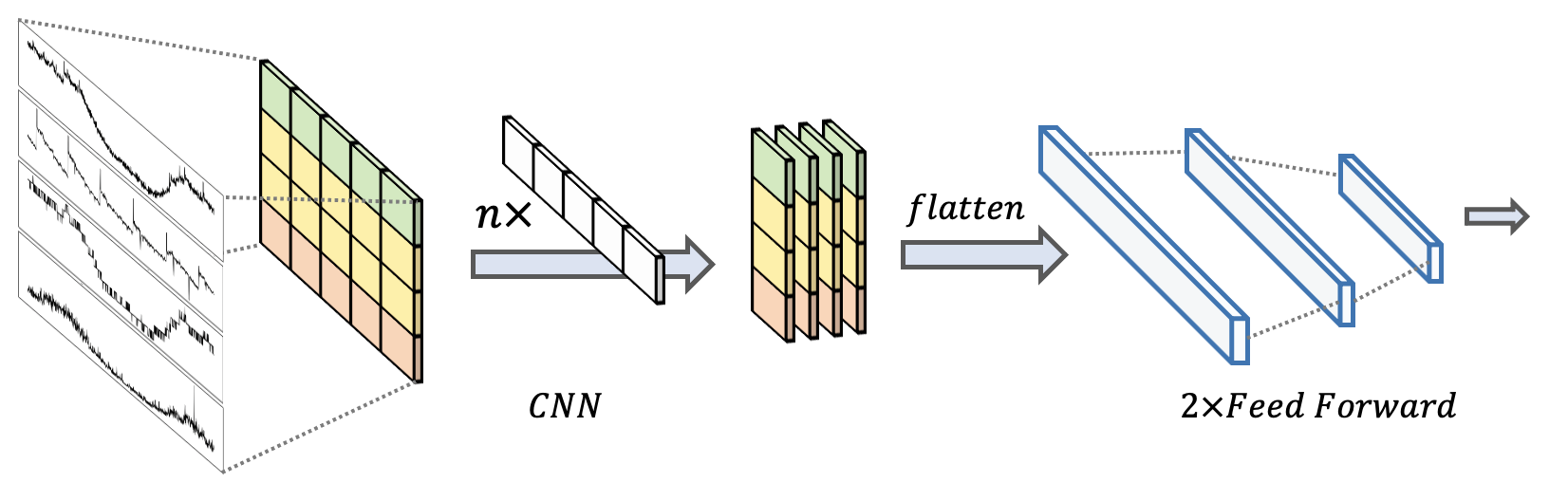}
    \caption{Expert Network.}
    \label{Fig.expert}
\end{figure}

In our implementation shown in Fig. \ref{Fig.expert}, the convolution layer contains $N$ kernels of width $l$ (equals to window size) and height 1. As a kernel sweeps through the input $\mathbf{w_t}$ including all $K$ metrics, each metric's window is convolved into a single value, so the layer produces a vector whose size is $N\times1\times K$. In this way, we obtain temporal dependency by the metric-level convolution operation, meanwhile obtaining inter-metric dependency by the shared filters. Then the vector is flattened and further fed into a two-layer fully-connected network, forming an embedding of the original input. All embeddings constitute a candidate set.

\subsection{Personalized \& Shared Gate}
\label{subsec: psgate}
A hybrid gate structure is employed to map the selected input to the weights of embeddings in the candidate set. If patterns of metrics differ greatly from each other, instead of sharing the same expert, the loss of backpropagation will induce them to take advantage of different experts, which is reflected by diverse weight combinations given by gates. This empowers the model to resolve conflicts and irrelevant drifts among metrics. Furthermore, we design a dual-gate mechanism. A shared gate $G_s$ receives all selected windows from input, yet a personalized gate $G_p^k$ belonging to the k-th metric only receives its own window. Since there are massive metrics, too many gates with the same weight make the backpropagation gradient more chaotic for experts. A shared gate with a greater weight can learn the robust mapping relationships between expert fusion and input from more data and is more likely to induce experts to converge on dominant characteristics. We combine shared gate and personalized gate to leverage the advantages of both, making personalized gate act as an auxiliary role to tune the subtle differences between metrics. The hybrid gate is formulated as follows:
\begin{equation}
G_s(\mathbf{w_t^k}) = W_s\mathbf{w_t^k},\quad\, G_p^k(\mathbf{w_t^k}) = W_{pk}\mathbf{w_t^k},    
\end{equation}

\begin{equation}
    G_k(\mathbf{w_t^k}) = softmax(\varepsilon \cdot G_s + (1-\varepsilon)\cdot G_p^k).
\end{equation}
where $W_s$ and $W_{pk}$ ($k=1,2,\cdots,K$) are trainable matrixs, and $\varepsilon > 0.5$ is a weight coefficient of the shared gate.
\subsection{Tower Network}
\label{subsec: tower}
In our framework, the purpose of the tower network is to condense the embeddings into one final predicted value. We utilize two dense layers along with the activation function $ReLU$ and a dropout layer to reduce the dimensions. All towers' predictions are appended to set $\mathbf{\hat{y_t}}$. As the ground truth observation at time $t$ is $\mathbf{y_t}$, the final training objective $\mathcal{L}$ aims to minimize the squared L2 norm:
\begin{equation}
    \mathcal{L} = ||\mathbf{y_t} - \mathbf{\hat{y_t}}||_2^2 = \frac{\sum_{k=1}^{K}(y_k^{(t)} - \hat{y}_k^{(t)})^2}{K}
\end{equation}
Here, $||.||_2$ denotes the L2 norm. $\mathcal{L}$ is also used in the inference phase as anomaly score, which is further compared to the threshold set in line with previous detection performance in online detection scenarios.

\subsection{Loss Balancing}
\label{subsec: balance}
Balancing the weight of loss for each metric is a crucial topic as improper weights could narrow the perspective of detection and neglect some important metrics. Thanks to the homogeneity among metrics and even between inputs and outputs in MTS anomaly detection, we are able to alleviate the problem of imbalance simply by proper data preprocessing. For datasets whose metrics are not on the same order of magnitude, we normalize the original training data via a MinMax Scaler. Hence the loss of metric won't be too far from each other. More details about data preprocessing can be found in Section \ref{sec: data}.

\section{EXPERIMENTAL SETUP}
\label{sec:settings}
In this section, we first introduce the experimental datasets and evaluation metrics that are widely used in MTS anomaly detection domain. Then we present the details of hyperparameters settings.

\subsection{Datasets and Evaluation Metrics}
\label{sec: data}
We conduct experiments based on three real-world MTS datasets to evaluate the effectiveness of CAD: SMD~\cite{omni}, SWaT~\cite{swat} and WADI~\cite{wadi}. All of these datasets are public and universally employed in previous works~\cite{interfusion,omni,tranad,thoc}. 
The summary of these datasets is listed in Table \ref{table.data}, including the number of entities, dataset size, the number of dimensions and anomaly ratio in the test set.
\begin{table}[htb]
\caption{Dataset Statistics.}
\begin{tabular}{cccccc}
\toprule
Dataset & Entities & Metrics & Train  & Test   & Anomaly (\%) \\ \hline
SMD     & 28       & 38         & 708405 & 708420 & 4.16        \\
SWaT    & 1        & 51         & 495000 & 449919 & 11.98       \\
WADI    & 1        & 123        & 784517 & 172801 & 5.77        \\ 
\bottomrule
\end{tabular}
\label{table.data}
\end{table}

As the range of each metric in SMD has been limited to 0-1, we skip the data preprocessing of this dataset. For SWaT and WADI whose readings range from $10^{-2}$ to $10^{3}$, we apply MinMax Scaler to limit the value of the training set to 0-1. The maximum and the minimum value of the training set are further used as criteria to normalize the test set. Time series are then clipped to a proper scale. 
More details about the datasets' particulars and preprocessing can be found in our repository.

In a real-time system, anomalies generated by system or external factors tend to persist for some time (e.g. bugs in programs cause sustained high CPU usage), forming a contiguous anomaly segment. Human operators rarely care about point-wise metrics in applications. Thus we apply the point-adjustment approach introduced by ~\cite{pa}, based on the assumption that it is reasonable to consider the anomaly segment has been detected if at least one moment within the segment triggers the alert. If any point within the contiguous anomaly segment from ground truth is marked as an anomaly, the whole segment is considered to be detected correctly. Additionally, in practice, an alert after a long delay is futile since the sooner the anomaly is identified, the less damage it causes to the system or the service. According to this presumption, we also adopt the k-th PA approach proposed by ~\cite{kthpa}, which assumes that the anomaly segment is recognized correctly only if the delay of the detected point is less than k from the start point of this segment.  

  We employ Precision (P), Recall (R) and F1-score to evaluate the performance of our method and baselines based on the above two approaches respectively. These metrics are calculated as follows:
\begin{displaymath}
    P = \frac{TP}{TP+FP},\quad R = \frac{TP}{TP+FN},\quad F1=\frac{2\cdot P \cdot R}{P+R}
\end{displaymath}

where TP denotes True Positives, FP denotes False Positives and FN denotes False Negatives. For each entity, we enumerate all possible anomaly thresholds to find the optimal one for getting the highest anomaly scores when calculating F1-score~\cite{madgan, usad, interfusion}. The result is denoted as \bm{$F1_{best}$}. We discard threshold selection methods like POT~\cite{omni} because they introduce parameters that need to be tuned, which is unfair for those methods that do not provide these parameters. The purpose of using \bm{$F1_{best}$} is not meant to find a threshold for performance evaluation, but to directly measure a model's optimal performance without introducing any hyperparameter, which ensures fair evaluations for baselines. Several datasets contain more than one MTS entity. For instance, SMD embodies time series from 28 machines distributed in three clusters. In this case, the F1-score represents the average of all machines’ F1-best. In ~\cite{omni}, the authors use the average precision (denoted \bm{$\bar{P}$}) and average recall (denoted \bm{$\bar{R}$}) to get the F1-score. This measure is denoted \bm{$F1^{*}$} score in our experiments. As \bm{$\bar{P}$} and \bm{$\bar{R}$} neutralize some severe deviations in original precisions and recalls, in the case of uneven data distribution among metrics, \bm{$F1^{*}$} score usually exceeds \bm{$F1_{best}$}.


\subsection{Hyperparameters Settings}
We select different hyperparameter combinations empirically on different datasets for CAD and its variant. Due to the differences in characteristics among the datasets, the window of time series is set to 16 for SMD, 32 for SWaT and WADI. The number of experts is 5 for SMD, 7 for WADI and 9 for SWaT. As a prediction-based method, we use horizon as one hyperparameter, which means the distance between the predicted timestamp and the last timestamp in the window. We set the horizon to 3 for SMD which is more stable, and 1 for SWaT and WADI. 

For all datasets, the number of kernels in one expert network is set to 16 and the weight coefficient of the shared gate $\varepsilon$ is set to 0.7 empirically. We apply the Adam optimizer and CosineLR scheduler with an initial learning rate of 0.001 to optimize models. Batch sizes are set to 128 in the training process. The early stopping strategy is adopted while the maximum epoch for training is set to 10. 

All variants in the ablation study use the same hyperparameters as CAD. The models of baselines are trained using the parameters provided in their article. Specifically, we train and test Anomaly Transformer on each entity in SMD instead of on concatenated one shown in their source code for a fair comparison.

\section{Evaluation}
\label{sec:evaluation}
We conduct quantitative and qualitative experiments to evaluate the performance of our model. Four research questions need to be answered urgently during the experiments:

\noindent {\bfseries RQ1:} How does our model perform on public datasets compared to other state-of-the-art approaches?

\noindent {\bfseries RQ2:} How much does each constituent in our design contribute to overall performance?

\noindent {\bfseries RQ3:} Is CAD robust enough under various parameter settings?

\noindent {\bfseries RQ4:} Can different experts learn the representations of data from different perspectives?

\subsection{Baseline Approaches}
To demonstrate the merit of our proposed algorithm, we select nine recent state-of-the-art unsupervised methods for multivariate time series anomaly detection for comparison with CAD, including both prediction-based and reconstruction-based approaches. All methods' source codes are available in Github\footnote{Source code of LSTM-NDT comes from \url{https://github.com/khundman/telemanom}. DAGMM, MAD-GAN, USAD and TranAD come from \url{https://github.com/imperial-qore/TranAD}. OmniAnomaly comes from \url{https://github.com/NetManAIOps/OmniAnomaly}. Interfusion comes from \url{https://github.com/zhhlee/InterFusion}. DVGCRN comes from \url{https://github.com/BoChenGroup/DVGCRN}.}. We use full-size datasets for all methods.

\begin{itemize}
    \item {\bfseries LSTM-NDT}~\cite{lstmndt} uses the LSTM-RNN model to attain high prediction accuracy and provides an unsupervised threshold selection method to dynamically evaluate residuals.
    \item {\bfseries DAGMM}~\cite{dagmm} learns a low-dimensional embedding of the original time series via a deep autoencoder (AE), then feeds the embedding and reconstruction error of AE into Gaussian Mixture Model to predict their likelihood.
    \item {\bfseries MAD-GAN}~\cite{madgan} employs LSTM-RNN as GAN’s base model. Contrasting with conventional GAN, MAD-GAN takes the whole variable set into account concurrently in order to draw the inter-metric dependency between metrics.
    \item {\bfseries OmniAnomaly}~\cite{omni} and {\bfseries Interfusion}~\cite{interfusion} are methods based on variational auto-encoder to denoise the anomalies and capture dependencies via hierarchical stochastic latent variables.
    \item {\bfseries USAD}~\cite{usad} and {\bfseries TranAD}~\cite{tranad} apply the idea of adversarial learning and design a two-stage training framework, combining the advantages of autoencoders/self-attention encoders and adversarial training.
    \item {\bfseries DVGCRN}~\cite{dvgcrn} adopts an adaptive variational graph convolutional recurrent network unit to capture spatial and temporal fine-grained correlations, which are further extended into a deep variational network.
    \item {\bfseries Anomaly Transformer}~\cite{anot} utilizes an attention mechanism to compute the association discrepancy and further amplifies it via a minimax strategy.
\end{itemize}

\begin{table*}[htb]
\caption{Performance comparison under the Point-Adjustment (PA) approach. Best scores are highlighted in bold, and second best scores are highlighted in bold and underlined.}
\begin{tabular}{c|cccc|ccc|ccc}
\toprule
\hline
\multirow{2}{*}{Method} & \multicolumn{4}{c|}{SMD}           & \multicolumn{3}{c|}{SWaT} & \multicolumn{3}{c}{WADI} \\ \cline{2-11} 
                        & $\bar{P}$      & $\bar{R}$     & $F1$     & $F1^*$    & $P$      & $R$     & $F1$     & $P$      & $R$     & $F1$          \\ \hline
LSTM-NDT                & 0.9033 & 0.8479 & 0.8623 & 0.8747 & 0.9750 & 0.7120 & 0.8230 & 0.5340  & 0.4564 & 0.4921 \\ \hline
DAGMM                         & 0.9130    & 0.9391    & 0.9163 & 0.9258 & 0.9797 & 0.7456 & 0.8468 & 0.8873 & 0.5111 & 0.6486 \\ \hline
MAD-GAN                       & 0.8601    & 0.7576    & 0.7712 & 0.8056 & 0.8868 & 0.7999 & 0.8411 & 0.8933 & 0.3842 & 0.5374 \\ \hline
USAD                          & 0.9269    & 0.9393    & 0.9228 & 0.9330 & 0.9365 & 0.8421 & 0.8868 & 0.4893 & 0.5606 & 0.5225 \\ \hline
OmniAnomaly                   & 0.9396    & 0.9344    & 0.9300 & 0.9370 & 0.8777 & 0.7884 & 0.8306 &  0.6360 & 0.5391 & 0.5836 \\ \hline
TranAD                        & 0.9228    & 0.9492    & 0.9244 & 0.9358 & 0.9566 & 0.7734 & 0.8553 & 0.9920 & 0.3728 & 0.5420 \\ \hline
InterFusion                   & 0.9396    & 0.9344    & 0.9300 & 0.9327 & 0.9858 & 0.8481 & 0.9118 & 0.9510 & 0.7683 & 0.8499 \\ \hline
DVGCRN                        & 0.9596    & 0.9604    & \underline{\textbf{0.9578}} & \underline{\textbf{0.9600}} & 0.6842 & 0.8674 & 0.7650 & 0.5779 & 0.6228 & 0.5995 \\ \hline
Anomaly Transformer           & 0.8999    & 0.8953    & 0.8875 & 0.8976 & 0.9760 & 0.8761 & \underline{\textbf{0.9234}} & 0.8226 & 1.0000 & \underline{\textbf{0.9026}} \\ \hline
\textbf{CAD} & 0.9624    & 0.9914    & \textbf{0.9760} & \textbf{0.9767} & 0.9542 & 0.9122 & \textbf{0.9327} & 0.9301 & 0.9124 & \textbf{0.9212} \\ \hline
\bottomrule
\end{tabular}
\label{table.sum}
\end{table*}
\begin{table*}[htb]
\caption{Performance comparison under the k-th PA approach on SMD. Delay denotes the detection deadline in each anomaly segment. Only when outliers are detected within delay can the detection be recognized as a valid one.}
\scalebox{0.9}{
\begin{tabular}{c|cccc|cccc|cccc}
\toprule
\hline
\multirow{2}{*}{Delay (k)} & \multicolumn{4}{c|}{10 points}           & \multicolumn{4}{c|}{20 points} & \multicolumn{4}{c}{30 points} \\ \cline{2-13} 
                        & $\bar{P}$      & $\bar{R}$     & $F1$     & $F1^*$    & $\bar{P}$      & $\bar{R}$     & $F1$     & $F1^*$  & $\bar{P}$      & $\bar{R}$     & $F1$     & $F1^*$ \\ \hline
LSTM-NDT                      & 0.5277    & 0.5645    & 0.4977 & 0.5455 & 0.6395 & 0.6194 & 0.5937 & 0.6293 & 0.6857 & 0.6532 & 0.6366 & 0.6691 \\ \hline
DAGMM                         & 0.7774    & 0.7418    & 0.7187 & 0.7592 & 0.8127 & 0.7829 & 0.7668 & 0.7975 & 0.8128 & 0.8259 & 0.7854 & 0.8193 \\ \hline
MAD-GAN                       & 0.5903    & 0.5379    & 0.4864 & 0.5629 & 0.7453 & 0.5677 & 0.5938 & 0.6445 & 0.7508 & 0.5982 & 0.6140 & 0.6659 \\ \hline
USAD                          & 0.7566    & 0.7197    & 0.6939 & 0.7377 & 0.7903 & 0.8020 & 0.7547 & 0.7961 & 0.8209 & 0.8207 & 0.7779 & 0.8208 \\ \hline
OmniAnomaly                   & 0.7710    & 0.7076    & 0.6996 & 0.7380 & 0.7803 & 0.7711 & 0.7369 & 0.7757 & 0.7920 & 0.8048 & 0.7624 & 0.7983 \\ \hline
TranAD                        & 0.7699    & 0.8266    & \underline{\textbf{0.7627}} & \underline{\textbf{0.7973}} & 0.7978 & 0.8583 & 0.7966 & \underline{\textbf{0.8269}} & 0.8180 & 0.8765 & 0.8268 & 0.8462 \\ \hline
InterFusion                   & 0.7618 & 0.7448 & 0.7200 & 0.7532 & 0.8235 & 0.7898 & 0.7784 & 0.8063 & 0.8365 & 0.8183 & 0.8026 & 0.8273 \\ \hline
DVGCRN                        & 0.7755    & 0.7626    & 0.7395 & 0.7690 & 0.8555 & 0.7940 & \underline{\textbf{0.7994}} & 0.8236 & 0.8820 & 0.8543 & \underline{\textbf{0.8470}} & \underline{\textbf{0.8679}} \\ \hline
Anomaly Transformer                        & 0.4980    & 0.4206    & 0.4086 & 0.4560 & 0.5110 & 0.5274 & 0.4708 & 0.5190 & 0.5983 & 0.5704 & 0.5270 & 0.5840 \\ \hline
\textbf{CAD} & 0.8452    & 0.7861    & \textbf{0.7894} & \textbf{0.8146} & 0.8850 & 0.8287 & \textbf{0.8439} & \textbf{0.8598} & 0.8871 & 0.9060 & \textbf{0.8822} & \textbf{0.8965} \\ \hline
\bottomrule
\end{tabular}
}
\label{table.kth}
\end{table*}

\captionsetup{margin=0em}
\subsection{RQ1. Evaluation Results and Analysis}
\label{sec: drift}
Anomaly detection performance of CAD as well as all baselines is listed in Table \ref{table.sum} under the point-adjustment (PA) approach and Table \ref{table.kth} under the k-th point-adjustment (k-th PA) approach. CAD presents superior performance over other methods in general, achieving improvements by 0.009 to 0.429 under the normal PA approach and up to 0.380 under the k-th PA approach.

As shown in the table, most of the models work better on SMD, which is collected from a scenario involving real-time server clusters. Anomalies are relatively arresting on several machines (simultaneous large spikes on several metrics), enhancing the overall performance, especially \bm{$F1^{*}$} mentioned above. For example, almost all methods are able to obtain F1-scores up to 0.95 when evaluating machine-1-1. Nonetheless, time series in certain machines are quite misleading, specifically, data fluctuation in temporal dimension and intricate relationships in inter-metric dimension, requiring the model to obtain a higher capability of discerning subtle differences between normal patterns and anomalies. Performance of models on these machines varies greatly, contributing to the major discrepancy in final results. On machine-1-8, CAD earns a score of 0.9782, while other methods' scores range from 0.5961 to 0.9780. In terms of the whole SMD dataset, CAD outperforms baselines by 1.9\% (DVGCRN) to 26.5\% (MAD-GAN), exhibiting a powerful capacity to deal with a variety of situations. In addition, most existing models obtain rather poor grades on high-noise datasets like SWaT and WADI. Comparatively, our methodology performs better in terms of F1 with a slight compromise in precision, as a priority should be given to higher recall to some extent in anomaly detection tasks~\cite{fusagn}. As most of the methods perform well on SMD, we further adopt the k-th point-adjustment evaluation which is more discriminating and more practical to real-world detection. As shown in Table \ref{table.kth}, we assign the value of 10, 20 and 30 to $k$ respectively. Despite the fact that performance drops dramatically for some methods, CAD achieves the highest score in each case while maintaining a score of 0.7894 when $k$ is 10 and 0.8822 when $k$ is 30, exhibiting the effectiveness of our method in practice.

Further analysis is conducted based on baseline scores. LSTM-NDT and DAGMM are two unsupervised methods employing observation at a specific moment in time. In comparison with methods taking a sequence of observations as input, both cases are weak at exploiting temporal relationships during highly correlated periods of time~\cite{usad}. Moreover, LSTM-NDT predicts values for each metric separately, further leading to the loss of information embedded in inter-metric dependency~\cite{interfusion}. As a typical generative model, OmniAnomaly adopts bits of stochastic variables to model data distribution in a sequence of observations, yet a limited number of stochastic variables cannot extract complicated characteristics of time series sufficiently. Interfusion partially solves this problem by two-view stochastic variables, introducing an extra dimension to expand the representation space. However, these variables, along with MCMC imputation in the inference phase, bring an increase in model instability. Several runs yield results with some deviations. What's more, methods employing LSTM or GRU structures tend to consume extremely long training time, which is inapplicable to real-world situations.

Compared with baselines, CAD maximizes the effectiveness of temporal and inter-metric interconnections with the assistance of its well-designed structures. Such phenomena are observed that considerable drifts exist in some metrics over a prolonged period of time, which are labeled as normal patterns in both training sets and test sets. Due to the fact that certain metrics' trends are unstable inherently, even reasonable fluctuations in them will have a substantial impact on other metrics when using existing models. This results in the model misinterpreting them as an inter-metric anomaly. Expert selection mechanism in CAD can effectively shield this irrelevant influence and smooth anomaly score on these time slices. As a result, CAD is more sensitive to genuine anomalies. Meanwhile, this mechanism resolves the issue of the unpredictability of some metrics to some extent that has been criticized in traditional prediction-based methods~\cite{lstmae}.
\begin{figure}[htbp]
    \centering
    \subfloat[Metrics Visualization. Representative metrics are plotted with their raw value. Anomalous segments at the metric level are highlighted in red.]{
    \includegraphics[width=0.4\textwidth]{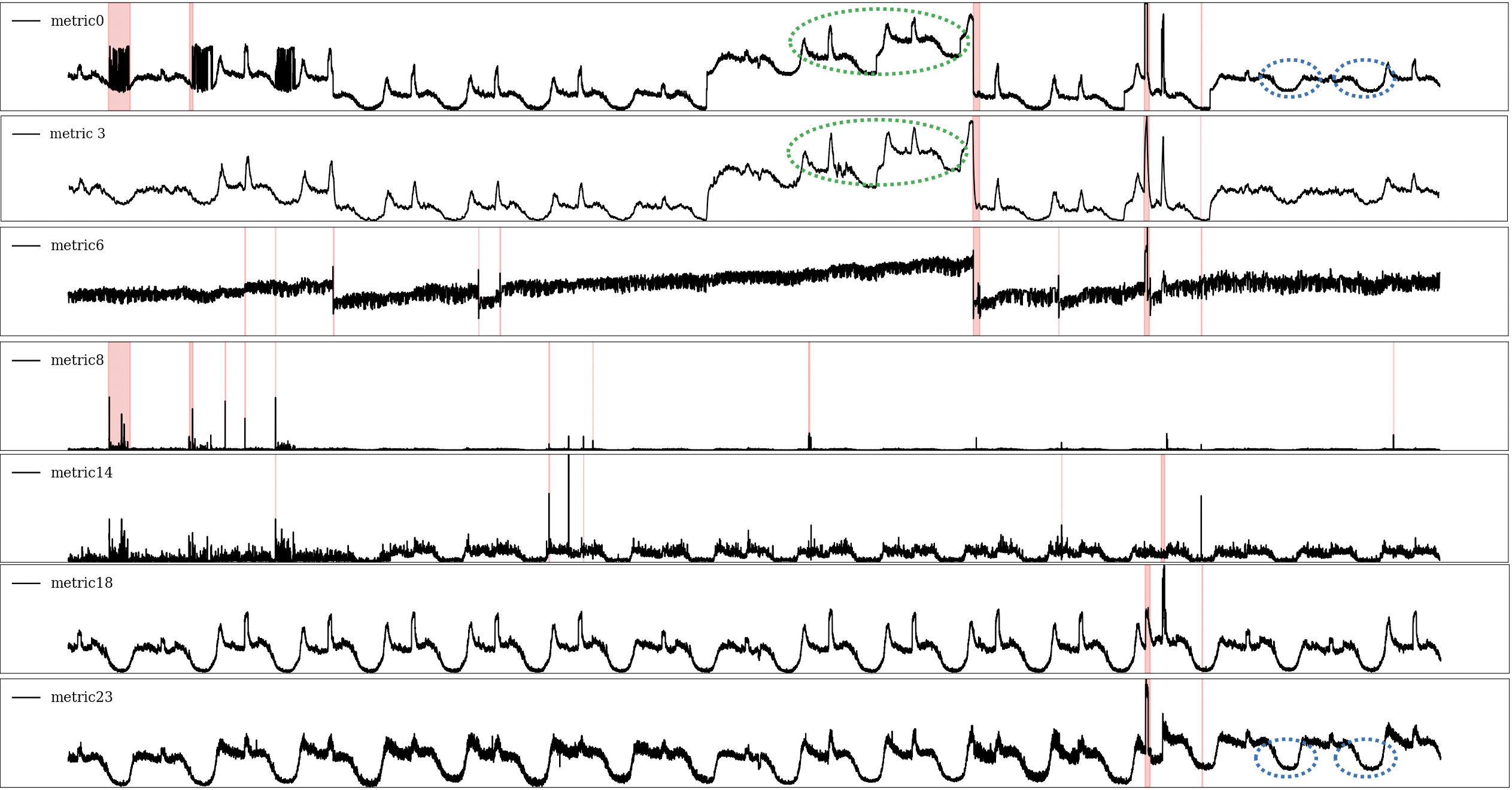}
    \label{fig.18gt}
    }
    \vspace{5mm}
    \\
    \subfloat[Anomaly scores of baselines. The best score threshold is represented by the red dotted line. Anomalous segments of all metrics are highlighted in red.]{
    \includegraphics[width=0.4\textwidth]{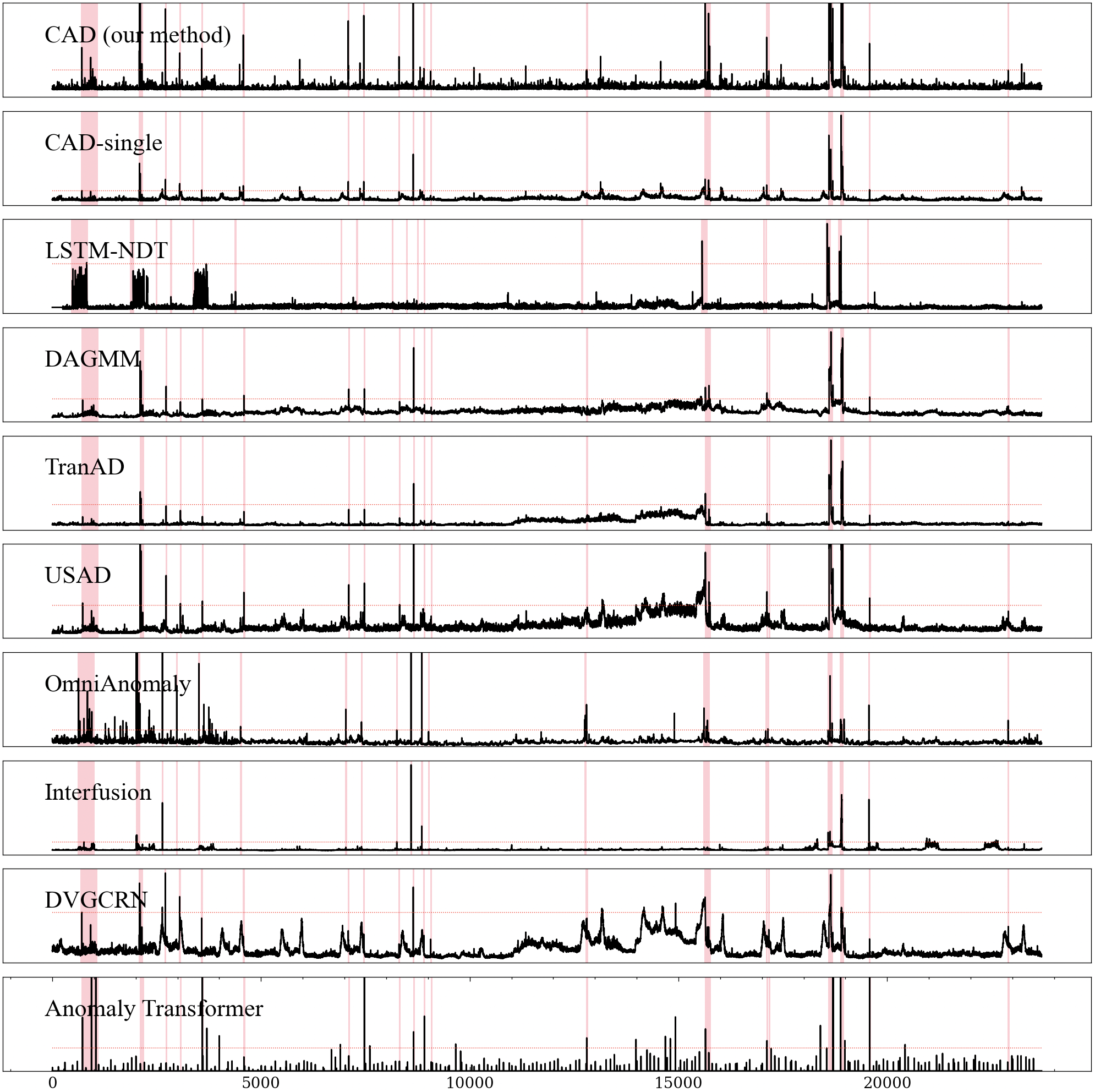}
    \label{fig.18}
    }
    \caption{Case study in machine-1-8.}
\end{figure}

\begin{figure}[htb]
	\centering
	\subfloat[CAD]{\includegraphics[width=.45\linewidth]{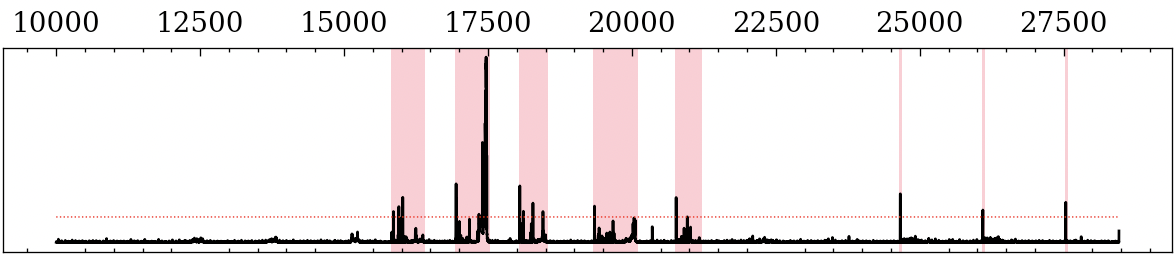}}\hspace{5pt}
	\subfloat[LSTM-NDT]{\includegraphics[width=.45\linewidth]{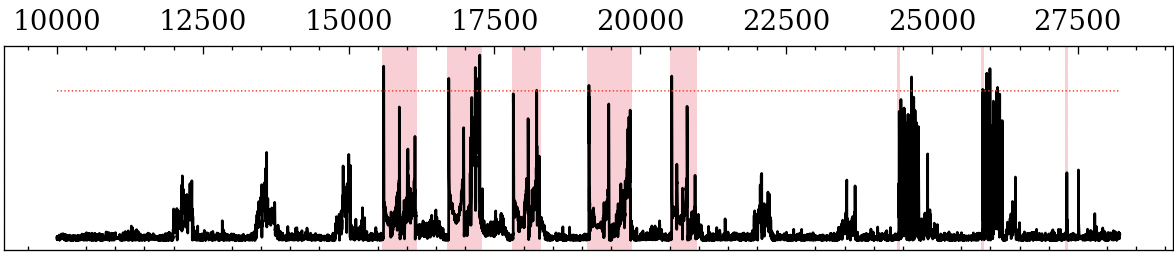}}\vspace{4pt}
	\subfloat[DAGMM]{\includegraphics[width=.45\linewidth]{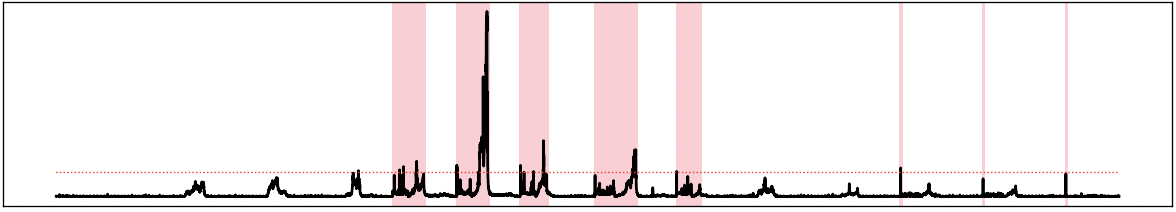}}\hspace{5pt}
	\subfloat[TranAD]{\includegraphics[width=.45\linewidth]{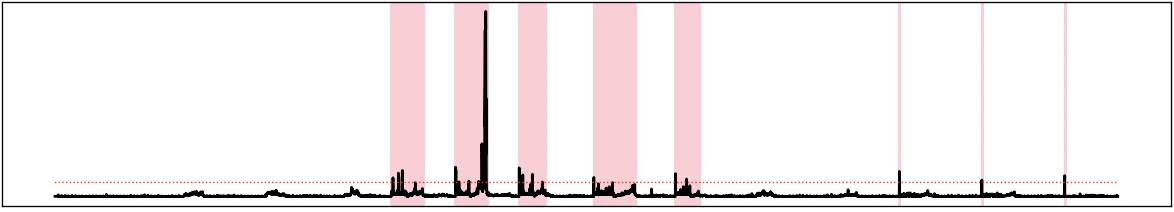}}\vspace{4pt}
    \subfloat[USAD]{\includegraphics[width=.45\linewidth]{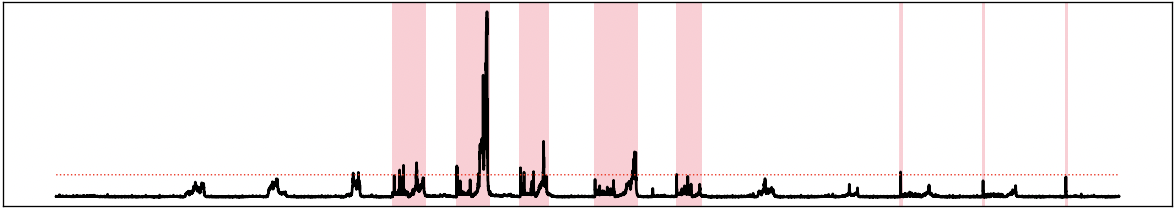}}\hspace{5pt}
	\subfloat[OmniAnomaly]{\includegraphics[width=.45\linewidth]{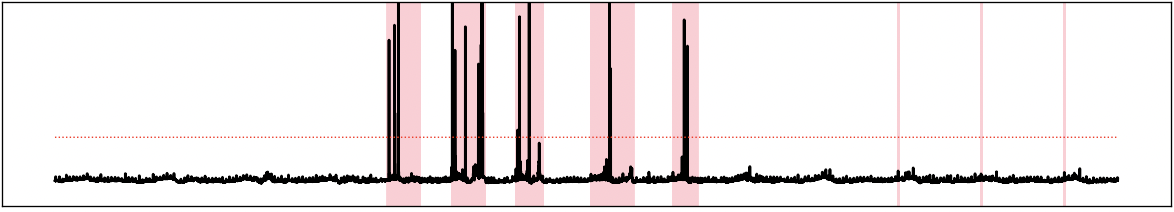}}\vspace{4pt}
	\subfloat[Interfusion]{\includegraphics[width=.45\linewidth]{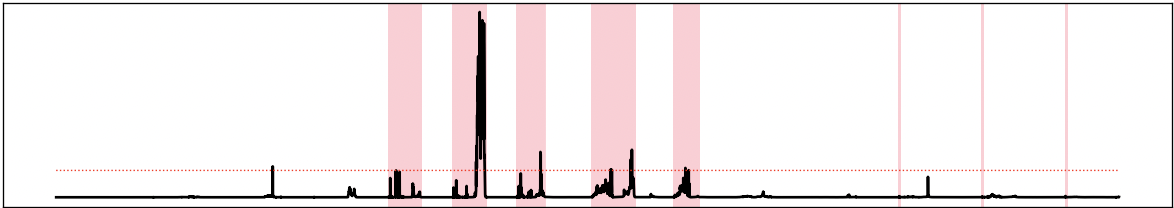}}\hspace{5pt}
	\subfloat[DVGCRN]{\includegraphics[width=.45\linewidth]{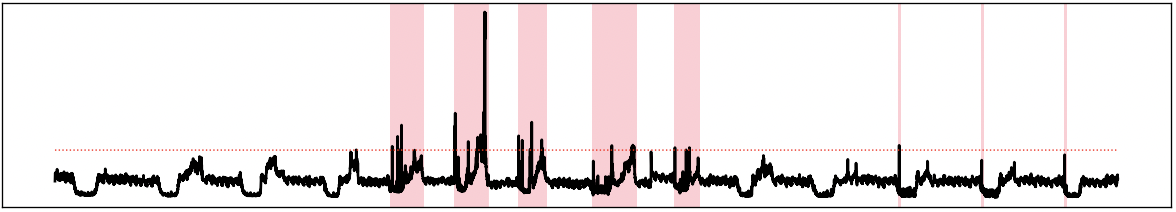}}
	\caption{Anomaly scores of baselines in machine-1-1 without first 10000 trivial (stable and normal) points. In all baselines, the best-F1 is greater than 0.95.}
    \label{fig.11}
\end{figure}

In detail, we conduct several case studies to illustrate the efficacy mentioned above. We visualize a few metrics in machine-1-8 included in SMD, on which models’ performance varies significantly, from 0.5761 to 0.9782 (Fig.~\ref{fig.18gt}). We also present anomaly scores achieved from models except for MAD-GAN which even does not converge on this dataset (Fig.~\ref{fig.18}). Despite the fact that both CAD and LSTM-NDT are forecasting-based methods, LSTM-NDT is too sensitive to fluctuation to distinguish anomalies from trivial noise. Anomaly scores of DAGMM and TranAD drift obviously on account of time series circled in green. These time series don’t violate data distribution but just get a higher baseline within reasonable limits as a matter of, e.g., load balancing strategy. Even though Interfusion handles the previous situation appropriately, it fails to earn proper scores circled in blue, which are even more "normal" than the previous one. We also plot the score of a variant of CAD, CAD-single, whose metrics are trained and tested separately. Due to the absence of inter-metric dependency, this variant works well on the segment mentioned above, yet performs poorly on other data. Instead, CAD works in both cases since it learns great lessons jointly from intra-metric and inter-metric dependency to find out the intrinsic normal pattern of each metric. The drifts are almost removed by the techniques in our framework according to the anomaly score of CAD. Even on datasets that all methods achieve high grades, as shown in Fig.~\ref{fig.11}, owing to its superior anti-interference capabilities, CAD exhibits more evident spikes when encountering anomalies.

\subsection{RQ2. Ablation Study}
\label{sec: ablation}
We omit every relevant component of the framework to observe the extent to which it impacts the performance of the model with reference to the F1-score on various datasets. By virtue of leveraging inter-metric dependency to jointly detect anomalies in time series with all metrics, it is indispensable to concentrate on two issues in terms of model framework design. First, we are interested in finding out if different experts can pick up specific traits of temporal patterns in distinct perspectives as expected, and if gates can automatically learn metric-specific combinations of representations generated by experts. Second, we want to know if dependency between metrics helps improve the model’s capability to catch anomalies, that is to say, if training with metrics of time series jointly outperforms the corresponding single-metric model. In the aspect of components in the model, are convolution units, which are designed for the time series task, able to effectively extract representations embedded in raw data? In addition, what is the relationship between the number of experts and the model’s performance?

\noindent {\bfseries Effectiveness of MoE framework.} We pick out a commonly used model in the field of multi-task learning, which is known as Shared-Bottom structure~\cite{MMOE}, as our baseline. Compared with CAD which has a group of bottom networks, it sends the shared representations to task-specific towers directly without going through gate networks, thus we refer to this structure  as {\itshape w/o gate}. To exclude effects induced by model complexity, for different numbers of experts, we modify the scale of bottom layers in the w/o gate structure such that the number of convolution kernels in the two models is equal. For example, the shared-bottom model with 80 kernels is contrasted with CAD which contains five experts, each of which has 16 kernels. The results of experiments suggest that the shared-bottom model has at most a 0.172 drop in performance in terms of F1-score, due to the fact that it copes with inherent conflict from irrelevant metrics badly. Sequentially, performance is rarely improved as the scale of the bottom layers grows. In comparison, CAD has awesome effectiveness owing to its flexible expert selection mechanism which handles this drawback gracefully. We also conduct a {\itshape single-task model} sharing the same structure as CAD yet each tower only has one exclusive expert network. Due to the inability to learn inter-metric dependency, it performs poorly even though the amount of parameters increases greatly.

\noindent {\bfseries Effectiveness of task-oriented feature selection.} Original time series contains multiple metrics, all of which do not have relations with a particular KPI. Instead of exposing all data to gate ({\itshape w/o selection}), the task-oriented selected window eliminates a lot of irrelevant information interference, which may confuse gates when computing weights. In our experiments, models without selection strategy have 0.0122 to 0.1625 drops on various datasets, demonstrating the great improvement brought by this mechanism. 

\noindent {\bfseries Effectiveness of personalized \& shared gate.} In general, CAD’s performance surpasses two models employing personalized gate ({\itshape w/o s-gate}) or shared gate ({\itshape w/o p-gate}) separately, especially on high dimensional datasets like SWaT. Due to the insufficient fineness of feature extraction, the model only using s-gate cannot achieve as high scores as the one using a hybrid gate. Without the robust s-gate, the model is more likely to overfit historical patterns. That’s the reason why shared gate performs even better than personalized gate which is considered to learn task-specific traits better in some cases. Experimental results show that the combination of the two gates further improves detection performance.

\noindent {\bfseries Effectiveness of convolution units.} In the model without convolution structure ({\itshape w/o conv}), experts consist of solely feed-forward networks. Temporal metrics in time windows are flattened as input to multi-layers neural networks. 
The absence of convolution units drastically degrades performance in general, and our framework is even rendered ineffective due to the lack of representations, demonstrating that convolution units along with two feed-forward layers in CAD have the ability to efficiently extract substantial temporal information embedded in raw time series.
\begin{table}[htb]
\caption{Performance of CAD and its variants in terms of best-F1 under point-adjustment.}
\centering
\setlength{\tabcolsep}{4.5mm}{
\begin{tabular}{cccc}
\toprule
Variants    & SMD    & SWaT   & WADI   \\ \hline
single-task & 0.9551 & 0.8739 & 0.8048 \\ 
w/o conv    & 0.9617 & 0.9101 & 0.7470 \\
w/o selection   & 0.9638 & 0.8899 & 0.7586 \\
w/o gate    & 0.9558 & 0.8488 & 0.7485 \\
w/o s-gate  & 0.9647 & 0.9271 & 0.8063 \\
w/o p-gate  & 0.9736 & 0.9223 & 0.8704 \\
\textbf{CAD}         & \textbf{0.9760} & \textbf{0.9327} & \textbf{0.9211}\\
\bottomrule
\end{tabular}
}
\label{table.abl}
\end{table}

\subsection{RQ3. Hyperparameter Sensitivity}
\begin{figure}[htb]
	\centering
	\subfloat[window size]{\includegraphics[width=.4\linewidth]{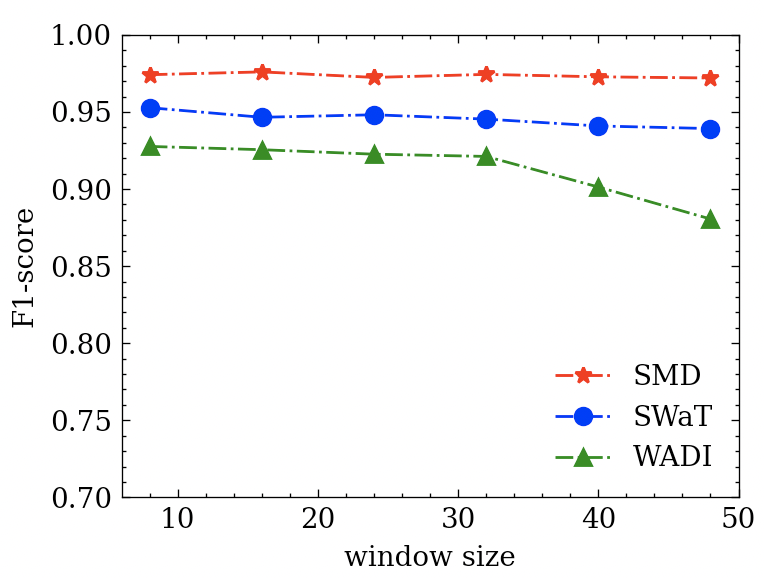}}\hspace{8pt}
	\subfloat[number of experts]{\includegraphics[width=.4\linewidth]{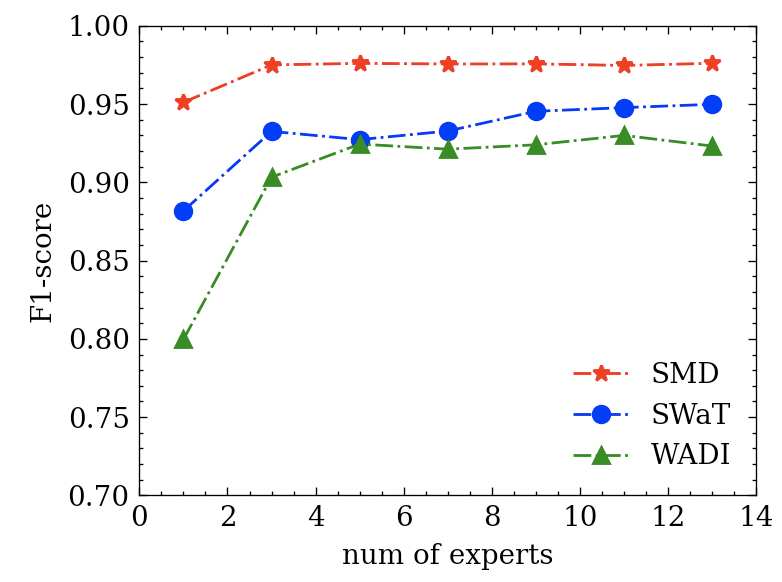}}
	\caption{Hyperparameter sensitivity under three datasets.}
    \label{fig.sens}
\end{figure}

As an application-oriented approach, the feasibility of a model deserves adequate consideration. We conduct contrast experiments in terms of key hyperparameters to test whether the model is parameter-sensitive (Fig. \ref{fig.sens}). One of these parameters is the window size. A larger window size means more long-term dependency within metrics, whereas introducing more remote time points that may dilute the significance of neighboring observations. In our experiments, the score of CAD has a slight downward trend as the window size exceeds 40. In general, however, CAD gets relatively smooth scores under a variety of settings, showing its robustness to window size. Another key parameter is the number of experts. When it is set to 1, CAD degrades to a shared-bottom model. With the increase in the number of experts, the model has a greater capacity to capture temporal and inter-metric dependency hidden in time series, while more parameters sacrifice the simplicity of the model, spending extra time training and testing data which is unacceptable for real-time anomaly detection. A value between 5 and 11 provides a reasonable trade-off between representation capacity and time efficiency. 

\subsection{RQ4. Feature Extraction Capability of Experts}
\begin{figure}[htbp]
    \centering
    \includegraphics[width=0.35\textwidth]{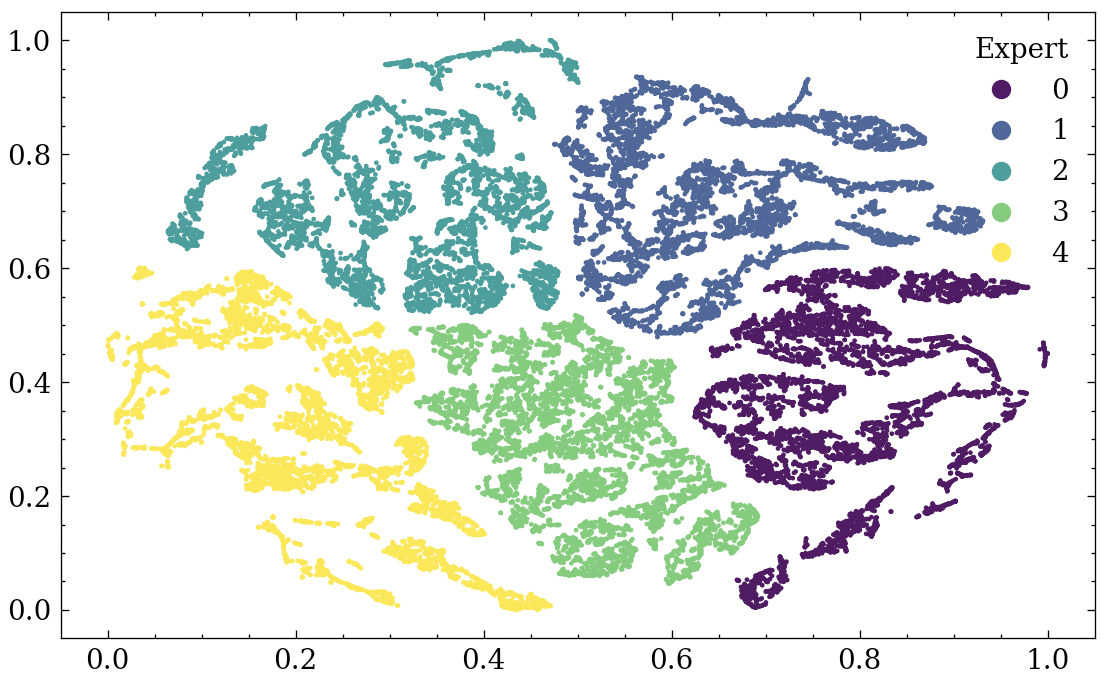}
    \caption{Visualization of embedding distributions of different experts. The high-dimensional vectors are mapped to a 2-dimensional space through t-SNE.}
    \label{Fig.tsne}
\end{figure}

A key prerequisite for modeling inter-metric promotion as well as conflict is that experts are able to learn the traits of time series from diverse perspectives. Given a time window as input, the expert network generates $E^{(t)}(\mathbf{w_t})$ whose shape is $N \times W$ where $N$ denotes the number of experts and $W$ denotes the dimension of the hidden vector. To analyze the validity of the expert network, we visualize the distribution of embeddings sampled from $E^{(t)}(\mathbf{w_t})$ during the test phase on machine-1-8, as the value of $N$ is 5 and $W$ is 128. Due to the lack of a straightforward way of viewing 128-dimensional space, we compress it to 2-dimensional space through t-distributed Stochastic Neighbor Embedding (t-SNE)~\cite{t-sne}. Each point in the scatter plot corresponds to a low-dimensional representation of the primitive hidden vector, meanwhile colored in line with the expert it comes from (Fig.~\ref{Fig.tsne}). In the t-SNE space, there are certain similarities among embeddings from the same expert since they aggregate together. On a global scale, the well-separated space further proves experts' ability to discern abundant yet distinct features.

\section{Related Work}
\label{sec:related work}
Deep neural networks have been proven to be highly effective at modeling intricate dependencies in time series over the past few years. As a result, they have become the method of choice in this field. Moreover, most proposed models are based on unsupervised learning due to the paucity of anomaly labels. Since anomalies are only small probability events, this approach works quite well. Taxonomically, these unsupervised solutions can be identified as reconstruction-based, forecasting-based and hybrid models~\cite{survey}.

\noindent \textbf{Reconstruction-based models.} These models encode subsequences of training data in latent space to filter out rare outlier points. Usually, a continuous sliding window is sent into the network and then mapped onto a low-dimension space. Soon a network expands the dimension of these data to reconstruct the input. In this process, an anomaly is more unlikely to be recovered as the model learns few abnormal patterns, resulting in a large gap between outputs and abnormal inputs, called reconstruction error, according to which the anomaly score is calculated. As an early attempt, EncDec-AD~\cite{lstmae} embeds encoder-decoder structure into the LSTM network to learn condensed temporal patterns. OmniAnomaly~\cite{omni} and Interfusion~\cite{interfusion} further employ stochastic variables to improve the robustness of the model. However, these methods are prone to error accumulation in long sequences~\cite{thoc}.

\noindent \textbf{Forecasting-based models.} The prerequisite of unsupervised forecasting-based models is that normal time series follows some rules, and anomalies are those who violate the inherent patterns. These models predict forthcoming points based on historical observations, then estimate if an anomaly occurs or not according to the point-wise difference between predicted values and ground truth values. Compared with traditional long sequence time-series forecasting tasks, anomaly detection asks for more precise prediction in a closer horizon and extended ability to cover diverse metrics concurrently. LSTM-NDT~\cite{lstmndt} is a well-known effort to handle this problem in a forecasting-based manner by using a non-parametric dynamic error thresholding strategy. THOC~\cite{thoc} uses a dilated skip-RNN structure to capture temporal dynamics and a hierarchical clustering process to fuse the multi-scale features. 

\noindent \textbf{Hybrid models.} These methods leverage composite errors, e.g., forecasting error or reconstruction error, to obtain the final anomaly score. USAD~\cite{usad}, as well as TranAD~\cite{tranad}, adopts two-phase training to amplify reconstruction errors. 
Generative Adversarial Network is another framework applied to reconstruct inputs. MAD-GAN~\cite{madgan} uses an RNN-based discriminator and generator to detect anomalies based on both reconstruction and discrimination losses. As a GNN-based method, FuSAGNet~\cite{fusagn} jointly optimizes reconstruction and forecasting errors. DVGCRN~\cite{dvgcrn} also computes both reconstruction and prediction scores via a graph convolutional recurrent network.

\section{Conclusion}
Today's software systems demand rapid responses to anomalies. Adequate analysis of MTS indicates that inappropriate use of inter-metric dependencies brings negative effects in some cases. In this paper, we propose Conflict-aware multivariate KPI Anomaly Detection (\textbf{CAD}), a novel 
unsupervised framework that effectively weeds out harmful conflicts which may confuse the detection for certain metrics. Based on the understanding of real-world data, we offer an exclusive structure for each metric to isolate the possible conflicts to a certain degree. What's more, the task-oriented feature-selection mechanism and a hybrid gate structure are elaborately designed to deal with the conflicts among metrics, greatly enhancing the effectiveness of the model. Time series-oriented experts learn rich characteristics from diverse perspectives under the combined effect of the above techniques.

We adopt a series of metrics for comprehensive evaluation. CAD outperforms all state-of-the-art baselines on three widely used datasets. Several experiments have demonstrated that CAD is adept in modeling intrinsic normal patterns while immune from irrelevant interference which results in false alarms or omissions in baseline methods. The concise structure and high computing efficiency allow it to be widely deployed in various scenarios and enable real-time detection. Moreover, the hyperparameter sensitivity study confirms its strong feasibility for a variety of detection tasks. 

\label{sec:conclusion}

\section*{ACKNOWLEDGMENTS}
This work was partially funded by the National Key Research and Development Program of China (No.2022YFB2901800), the National Natural Science Foundation of China (No.62202445 and No.72104229), the State Key Program of National Natural Science of China under Grant 62072264 and the CAS Program for fostering international mega-science (No.241711KYSB20200023).

\balance
\bibliographystyle{ACM-Reference-Format}
\bibliography{main}

\end{document}